\documentclass[review,11pt]{ReportTemplate}
\usepackage{times}
\usepackage{helvet}
\usepackage{courier}
\usepackage{graphicx}
\usepackage{epsf}
\usepackage{subfigure}
\usepackage{amsmath,amssymb}
\usepackage{xcolor}
\usepackage{paralist,algorithmic,algorithm}   % Compact enums: "compactenum", "compactitem"
\usepackage{url}    % Better email address and url handling

\def\sgn{\textrm{sgn}}

\def\l{\lambda}

\def\DD{{\cal D}}
\def\LL{{\cal L}}
\def\UU{{\cal U}}

\def\g{\gamma}

\def\bD{{\bf D}}
\def\bF{{\bf F}}

\def\L{\mbox{{\boldmath $\Lambda$}}}

\def\l{\lambda}

\def\h{{\bf h}}

\def\W{{\bf W}}

\def\x{{\bf x}}

\def\y{{\bf y}}
\def\0{{\bf 0}}
\def\1{{\bf 1}}

\def\x{{\bf x}}
\def\0{{\bf 0}}
\def\1{{\bf 1}}

\def\x{{\bf x}}

\def\C{{\cal C}}

\def\x{{\bf x}}

\begin{document}
\begin{frontmatter}

\title{Improving Semi-Supervised Support Vector Machines\\ Through Unlabeled Instances Selection}
\author{Yu-Feng Li}
\author{Zhi-Hua Zhou\corref{cor1}}
\address{National Key Laboratory for Novel Software Technology\\
Nanjing University, Nanjing 210093, China} \cortext[cor1]{\small Corresponding author. Email:
zhouzh@nju.edu.cn}

\begin{abstract}
Semi-supervised support vector machines (S3VMs) are a kind of popular approaches which try to
improve learning performance by exploiting unlabeled data. Though S3VMs have been found helpful in
many situations, they may degenerate performance and the resultant generalization ability may be
even worse than using the labeled data only. In this paper, we try to reduce the chance of
performance degeneration of S3VMs. Our basic idea is that, rather than exploiting all unlabeled
data, the unlabeled instances should be selected such that only the ones which are very likely to
be helpful are exploited, while some highly risky unlabeled instances are avoided. We propose the
S3VM-\emph{us} method by using hierarchical clustering to select the unlabeled instances.
Experiments on a broad range of data sets over eighty-eight different settings show that the chance
of performance degeneration of S3VM-\emph{us} is much smaller than that of existing S3VMs.
\end{abstract}


\end{frontmatter}

%%%%%%%%%%%%%%%%%%%%%%%%%%%%%%%%%%%%%%%%%%%%%%%%%%%%%%%%%%%%%%%%%%%%%%%%

\section{Introduction}

In many real situations there are plentiful unlabeled training data while the acquisition of class labels is costly and difficult. Semi-supervised learning tries to exploit unlabeled data to help improve learning performance, particularly when there are limited labeled training examples. During the past decade, semi-supervised learning has received significant attention and many approaches have been developed \cite{chapelle2006ssl,zhu2007semi,Zhou:Li2010}.

Among the many semi-supervised learning approaches, S3VMs (semi-supervised support vector machines) \cite{bennett1999sss,Joachims1999} are popular and have solid theoretical foundation. However, though the performances of S3VMs are promising in many tasks, it has been found that there are cases where, by using unlabeled data, the performances of S3VMs are even worse than SVMs simply using the labeled data \cite{zhang2000,chapelle2006ssl,Chapelle2008}. To enable S3VMs to be accepted by more users in more application areas, it is desirable to reduce the chances of performance degeneration by using unlabeled data.

In this paper, we focus on transductive learning and present the
S3VM-\emph{us} (S3VM with Unlabeled instances Selection) method. Our
basic idea is that, given a set of unlabeled data, it may be not
adequate to use all of them without any sanity check; instead, it
may be better to use only the unlabeled instances which are very
likely to be helpful while avoiding unlabeled instances which are
with high risk. To exclude highly risky unlabeled instances, we
first introduce two baselines, where the first baseline uses
standard clustering technique motivated by the discernibility of
density set \cite{SinghNIPS2008} while the other one uses label
propagation technique motivated by confidence estimation. Then,
based on the analysis of the deficiencies of the two baseline
approaches, we propose the S3VM-\emph{us} method, which employs
hierarchical clustering to help select unlabeled instances.
Comprehensive experiments on a broad range of data sets over
eighty-eight different settings show that, the chance of performance
degeneration of S3VM-\emph{us} is much smaller than that of TSVM
\cite{Joachims1999}, while the overall performance of S3VM-\emph{us}
is competitive with TSVM.

The rest of this paper is organized as follows. Section 2 briefly reviews some related work.
Section 3 introduces two baseline approaches. Section 4 presents our S3VM-\emph{us} method.
Experimental results are reported in Section 5. The last section concludes this paper.

%%%%%%%%%%%%%%%%%%%%%%%%%%%%%%%%%%%%%%%%%%%%%%%%%%%%%%%%%%%%%%%%%%%%%%%%

\section{Related Work}\label{sec:review+analysis}

Roughly speaking, existing semi-supervised learning approaches
mainly fall into four categories. The first category is generative
methods, e.g., \cite{Miller:Uyar1997,nigam2000text}, which extend
supervised generative models by exploiting unlabeled data in
parameter estimation and label estimation using techniques such as
the EM method. The second category is graph-based methods, e.g.,
\cite{blum2001,zhu2003,zhou2004}, which encode both the labeled and
unlabeled instances in a graph and then perform label propagation on
the graph. The third category is disagreement-based methods, e.g.,
\cite{blum1998,zhou2005tri}, which employ multiple learners and
improve the learners through labeling the unlabeled data based on
the exploitation of disagreement among the learners. The fourth
category is S3VMs, e.g., \cite{bennett1999sss,Joachims1999}, which
use unlabeled data to regularize the decision boundary to go through
low density regions \cite{Chapelle2005}.

Though semi-supervised learning approaches have shown promising performances in many situations, it has been indicated by many authors that using unlabeled data may hurt the performance
\cite{nigam2000text,zhang2000,Cozman2003,zhou2005tri,chawla2005learning,LaffertyNIPS2007,ben2008does,SinghNIPS2008}.
In some application areas, especially the ones which require high reliability, users might be reluctant to use semi-supervised learning approaches due to the worry of obtaining a performance worse than simply neglecting unlabeled data. As typical semi-supervised learning approaches, S3VMs also suffer from this deficiency.

The usefulness of unlabeled data has been discussed theoretically
\cite{LaffertyNIPS2007,ben2008does,SinghNIPS2008} and validated empirically
\cite{chawla2005learning}. Many literatures indicated that unlabeled data should be used carefully.
For generative methods, Cozman et al. \cite{Cozman2003} showed that unlabeled data can increase
error even in situations where additional labeled data would decrease error. One main conjecture on
the performance degeneration is attributed to the difficulties of making a right model assumption
which prevents the performance from degenerated by fitting with unlabeled data. For graph-based
methods, more and more researchers recognize that graph construction is more crucial than how the
labels are propagated, and some attempts have been devoted to using domain knowledge or
constructing robust graphs \cite{balcan2005person,Jebara2009}. As for disagreement-based method,
the generalization ability has been studied with plentiful theoretical results based on different
assumptions \cite{blum1998,dasgupta2002pac,wang2007,Wang:Zhou2010}. As for S3VMs, the correctness
of the S3VM objective has been studied on small data sets \cite{Chapelle2008}.

It is noteworthy that though there are many work devoted to cope with the high complexity of S3VMs \cite{Joachims1999,collobert2006lst,Chapelle2008,li2009means3vm}, there was no proposal on how to reduce the chance of performance degeneration by using unlabeled data. There was a relevant work which uses data editing techniques in semi-supervised learning \cite{li2005setred}; however, it tries to remove or fix suspicious unlabeled data during training process, while our proposal tries to select unlabeled instances for S3VM and SVM predictions after the S3VM and SVM have already been trained.

%By using domain knowledge, it is possible to construct a good graph which leads to good performance
%\cite{balcan2005person}. However, how to develop a method which works well in general cases remains
%an open problem. One interesting observation by \cite{zhu2005thesis} is weighted \emph{k}NN graphs
%with a small $k$ tend to perform well empirically. Recently, Jebara et.al.,(2009) indicate that
%$b$-matching graph, where each node owns the same number of edges, performs robustly better than
%\emph{k}NN graph.

\section{Two Baseline Approaches}\label{sec:algo}

As mentioned, our intuition is to use only the unlabeled data which are very likely to help improve the performance and keep the unlabeled data which are with high risk to be unexploited. In this way, the chance of performance degeneration may be significantly reduced. Current S3VMs can be regarded as an extreme case which believes that all unlabeled data are with low risk and therefore all of them should be used; while inductive SVMs which use labeled data only can be
regarded as another extreme case which believes that all the unlabeled data are high risky and therefore only labeled data are used.

%This idea relates to a few of SSL techniques considering unequal importance for different unlabeled
%instances which are proposed for different scenarios where the distributions of labeled and
%unlabeled data are different \cite{huangNIPS2006} or the proportions of different classes are
%imbalanced \cite{banasik2003sample}. Unlike above approaches which work as data pre-processing, we
%consider in this paper an opposite way, i.e., prediction post-processing. That is once you obtain
%the predicted results of S3VM, how to remove those risky prediction of S3VM such that the final
%performance could often better than that of supervised SVM using label data only while in the worst
%case, the performance could still be comparable to supervised SVM?

Specifically, we consider the following problem: Once we have obtained the predictions of inductive SVM and S3VM, how to remove risky predictions of S3VM such that the resultant performance could be often better and rarely worse than that of inductive SVM?

There are two simple ideas that are easy to be worked out to address the above problem, leading to two baseline approaches, namely S3VM-\emph{c} and S3VM-\emph{p}.

In the sequel, suppose we are given a training data set $\DD = \LL \bigcup \UU$ where $\LL =
\{(\x_1,y_1),\ldots,(\x_l,y_l)\}$ denotes the set of labeled data and $\UU =
\{\x_{l+1},\ldots,\x_{l+u}\}$ denotes the set of unlabeled data. Here $\x \in \mathcal{X}$ is an
instance and $y \in \{+1,-1\}$ is the label. We further let $y_{SVM}(\x)$ and $y_{S3VM}(\x)$ denote the
predicted labels on $\x$ by inductive SVM and S3VM, respectively.

%\begin{figure*}[htbp]
%\centering
%\begin{minipage}[c]{1.5 in}
%\centering
%\includegraphics[width =1.5 in]{pic1.eps}\;\;\;\;\;\;\;\;\;\;\;\;\;\;\;
%\centering \mbox{\footnotesize (a)}
%\end{minipage}
%\centering
%\begin{minipage}[c]{1.5 in}
%\centering
%\includegraphics[width =1.5 in]{pic2.eps}\;\;\;\;\;\;\;\;\;\;\;\;\;\;\;
%\centering \mbox{\footnotesize (b)}
%\end{minipage}
%\begin{minipage}[c]{1.5 in}
%\centering
%\includegraphics[width =1.5 in]{pic3.eps}
%\centering \mbox{\footnotesize (c)}
%\end{minipage}\\
%\begin{minipage}[c]{1.5 in}
%\centering
%\includegraphics[width =1.5 in]{pic4.eps}\;\;\;\;\;\;\;\;\;\;\;\;\;\;\;
%\centering \mbox{\footnotesize (d)}
%\end{minipage}
%\begin{minipage}[c]{1.5 in}
%\centering
%\includegraphics[width =1.5 in]{pic5.eps}\;\;\;\;\;\;\;\;\;\;\;\;\;\;\;
%\centering \mbox{\footnotesize (e)}
%\end{minipage}
%\begin{minipage}[c]{1.5in}
%\centering
%\includegraphics[width =1.5 in]{pic6.eps}
%\centering \mbox{\footnotesize (f)}
%\end{minipage}
%\caption{A demonstration with artificial three-moon data set. (a) Certain initial label setting.
%The two-color data points present the labeled examples for different classes and the grey data
%points present the unlabeled data. The red number in the block presents a group of four unlabeled
%instances. The rest of subfigures depict the results from (b) SVM with labeled data only; (c) S3VM;
%(d) S3VM-\emph{c} where each circle presents a cluster; (e) S3VM-\emph{p} and (f) Our method S3VM-\emph{us}.
%}\label{fig:illustration}
%\end{figure*}

\subsection{S3VM-\emph{c}} \label{sec:cluster}

The first baseline approach is motivated by the analysis in \cite{SinghNIPS2008} which suggests that unlabeled
data help when the component density sets are discernable. Here, one can simulate the component density sets by clusters and discernibility by a condition of disagreements between S3VM and inductive SVM. We consider the disagreement using two factors, i.e., \emph{bias} and \emph{confidence}. When S3VM obtains the same bias as inductive SVM and enhances the
confidence of inductive SVM, one should use the results of S3VM; otherwise it may be risky if we totally trust the prediction of S3VM.

\begin{figure*}[htbp]
\centering
\begin{minipage}[c]{1.6 in}
\centering
\includegraphics[width =1.6 in]{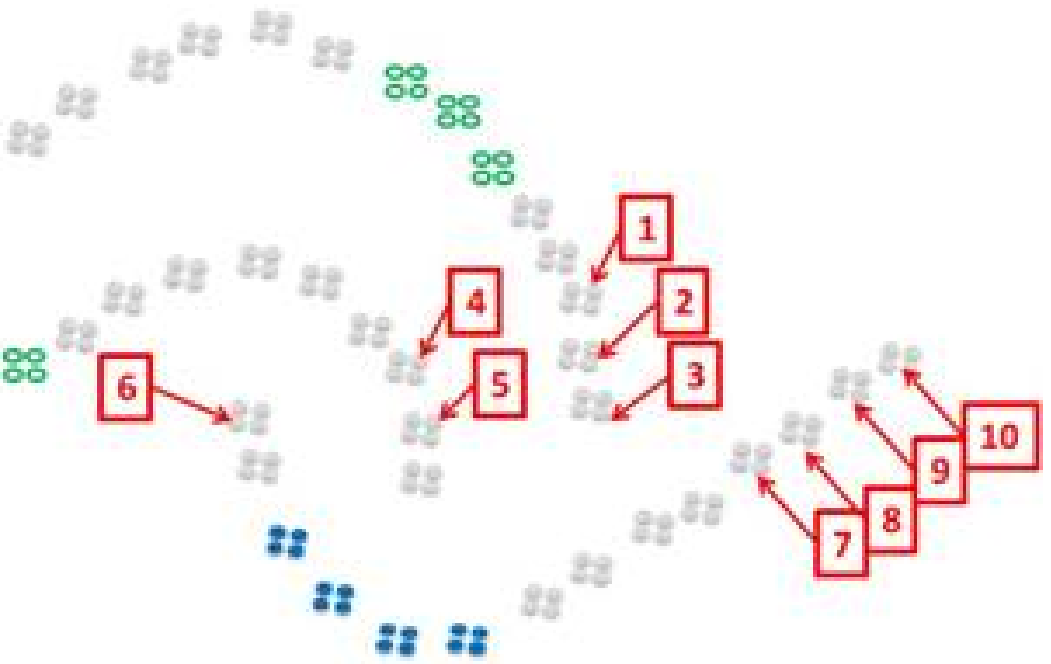}
\centering \mbox{\footnotesize (a)}
\end{minipage}
\centering
\begin{minipage}[c]{1.6 in}
\centering
\includegraphics[width =1.6 in]{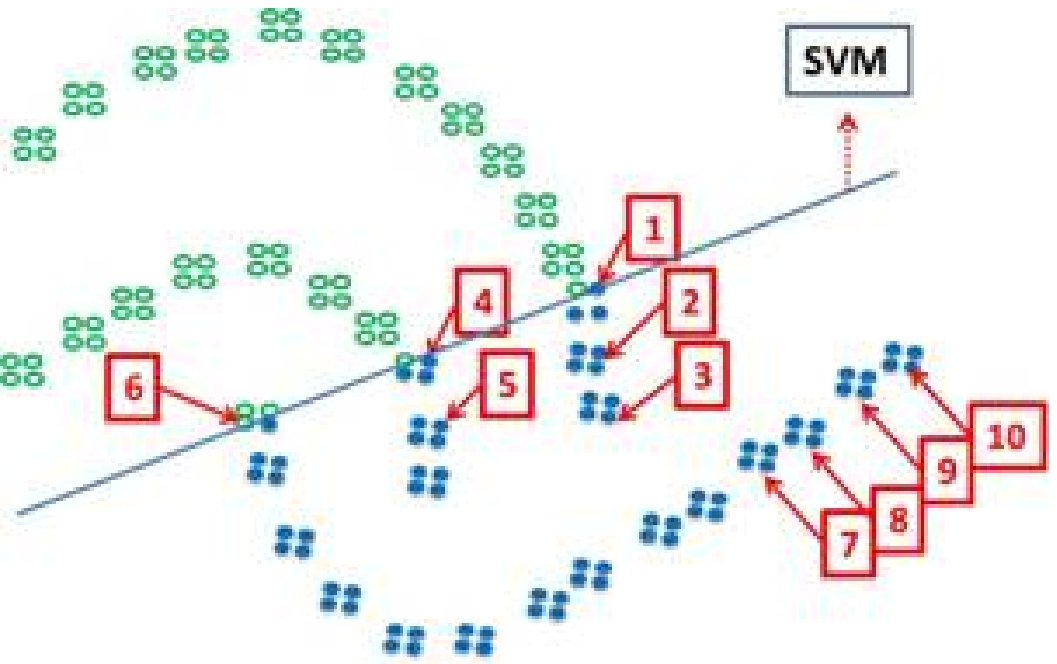}
\centering \mbox{\footnotesize (b)}
\end{minipage}
\begin{minipage}[c]{1.6 in}
\centering
\includegraphics[width =1.6 in]{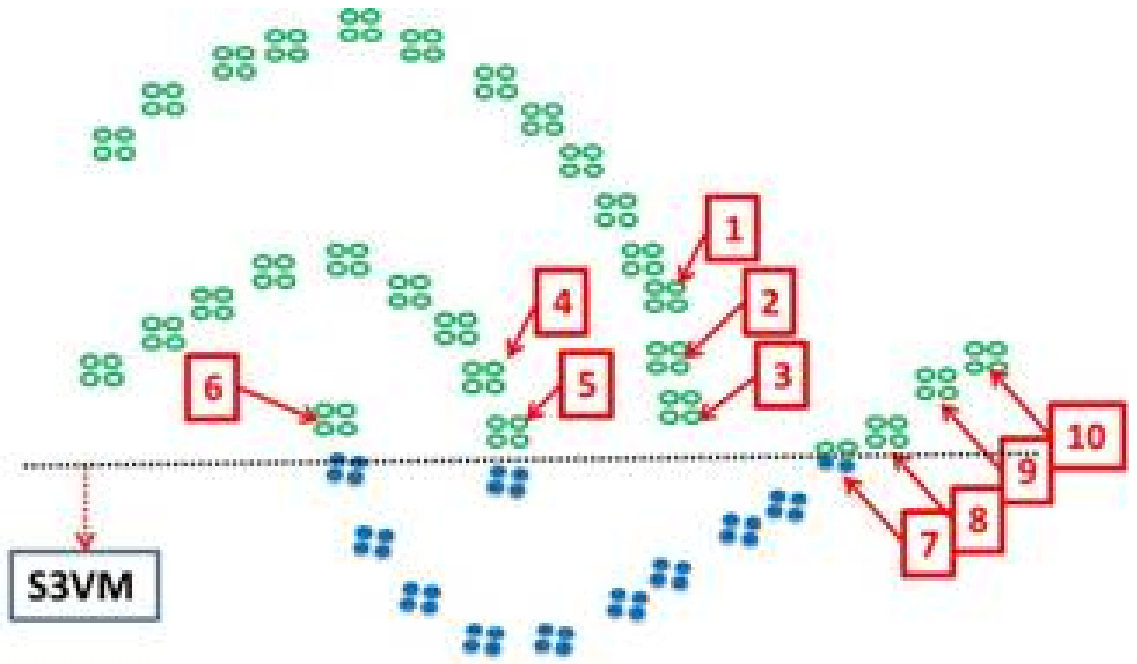}
\centering \mbox{\footnotesize (c)}
\end{minipage} \\
\begin{minipage}[c]{1.6 in}
\centering
\includegraphics[width =1.6 in]{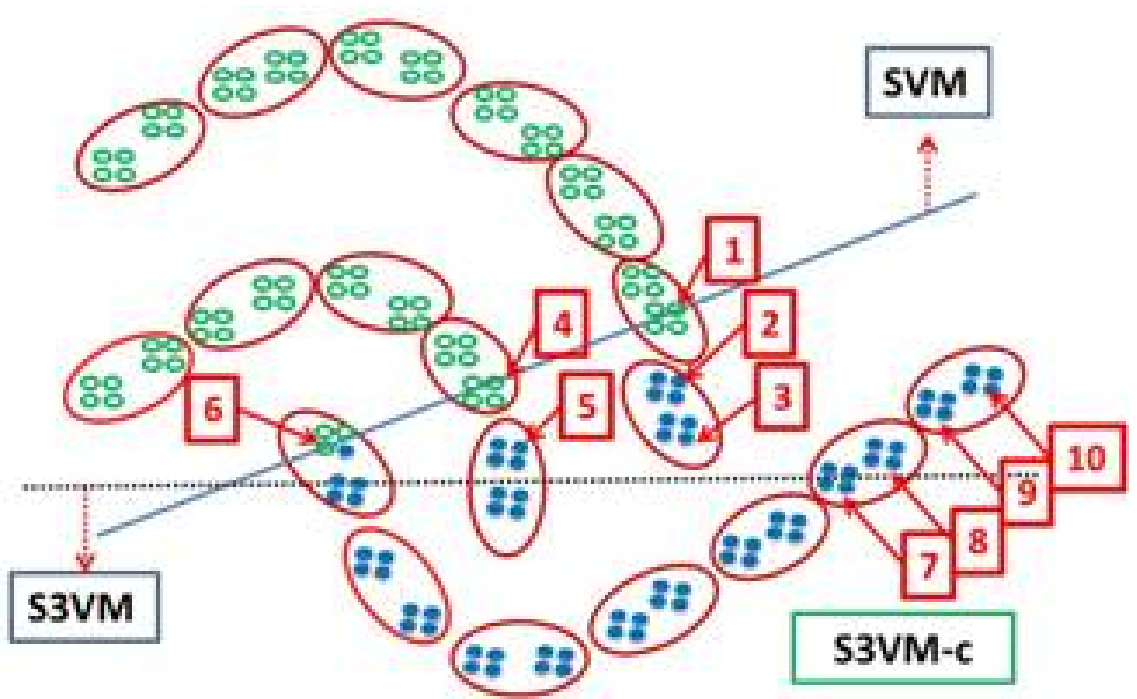}
\centering \mbox{\footnotesize (d)}
\end{minipage}
\begin{minipage}[c]{1.6 in}
\centering
\includegraphics[width =1.6 in]{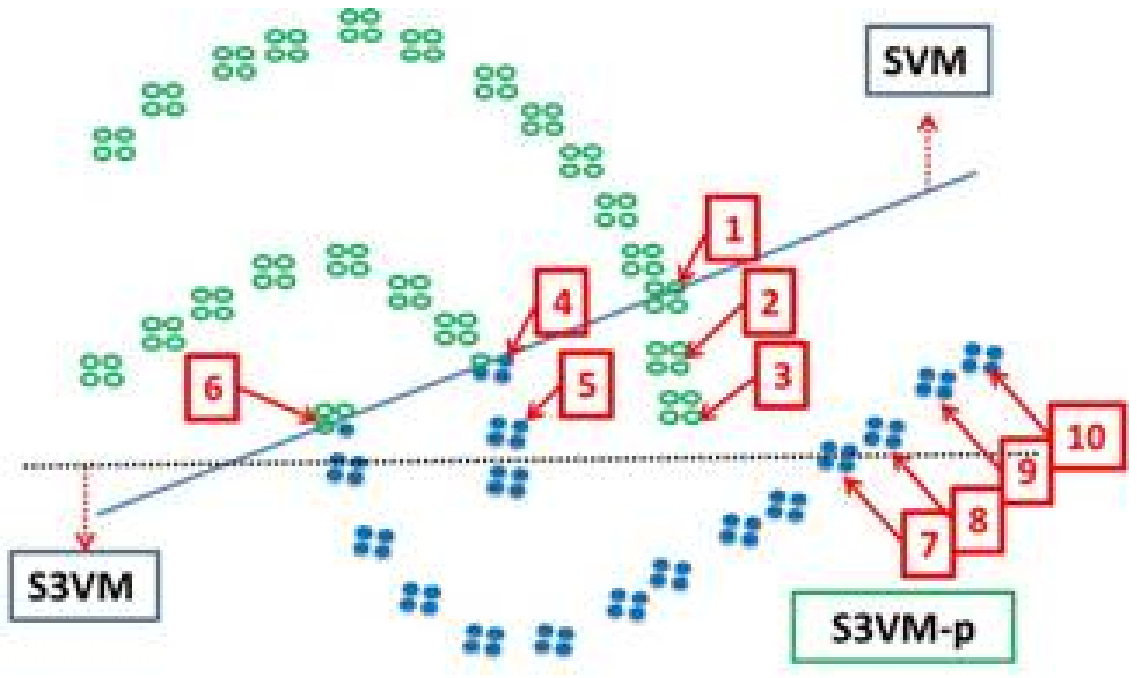}
\centering \mbox{\footnotesize (e)}
\end{minipage}
\begin{minipage}[c]{1.6 in}
\centering
\includegraphics[width =1.6 in]{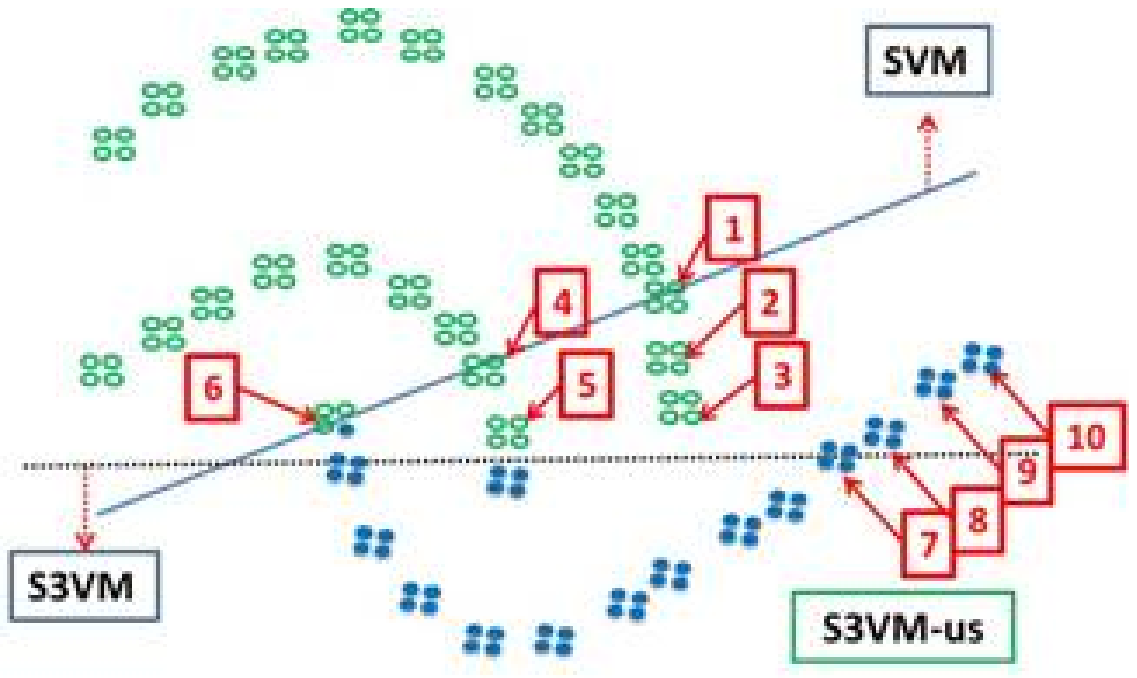}
\centering \mbox{\footnotesize (f)}
\end{minipage}
\caption{\small{Illustration with artificial three-moon data. (a) Labeled data (empty and filled
circles) and unlabeled data (gray points). The blocked numbers highlight groups of four unlabeled
instances. Classification results of (b) Inductive SVM (using labeled data only); (c) S3VM; (d)
S3VM-\emph{c}, where each circle presents a cluster; (e) S3VM-\emph{p}; (f) Our proposed
S3VM-\emph{us}.}}\label{fig:illustration}
\end{figure*}

Algorithm~\ref{alg:cluster} gives the S3VM-\emph{c} method and Figure~\ref{fig:illustration}(d) illustrates the intuition of S3VM-\emph{c}. As can be seen, S3VM-\emph{c} inherits the correct predictions of S3VM on groups $\{1,4\}$ while avoids the wrong predictions of S3VM on groups $\{7,8,9,10\}$.

%%%%%%%%%%%%%%%
% What is main motivation of using cluster assumption? Intuitive interpretation by geometric.
% How to use cluster assumption to design the concrete algorithm? Add some analysis about this method
%%%%%%%%%%%%%%%

\begin{algorithm}[h]\small
\caption{~~S3VM-\emph{c}} \textbf{Input}: $y_{SVM}$, $y_{S3VM}$, $\DD$ and parameter $k$
\begin{algorithmic}[1]
\STATE Perform partitional clustering (e.g., $k$means) on $\DD$. Denote
$\C_{1},\ldots,\C_{k}$ as the data indices of each cluster respectively.

\STATE For each cluster $i=1,\ldots,k$, calculate the label bias $lb$ and confidence $cf$ of SVM and S3VM according to:
\begin{eqnarray}\small
lb^i_{S(3)VM} &=& sign\left(\sum_{j \in \C_i} y_{S(3)VM}\left(\x_j\right)\right)
\nonumber \\
cf^i_{S(3)VM} &=& \left|\sum_{j \in \C_i} y_{S(3)VM}\left(\x_j\right)\right|. \nonumber
\end{eqnarray}
\STATE If $lb^i_{SVM} = lb^i_{S3VM} \;\;\& \;\; cf^i_{S3VM} > cf^i_{SVM}$, use the prediction of S3VM;
otherwise use the prediction of SVM.
\end{algorithmic}\label{alg:cluster}
\end{algorithm}

%%%%%%%%%%%%%%%%%
% What is main motivation of using label propagation? Find the uncertain/unconfident label assignment.
% How to use label propagation to design the concrete algorithm? Add some analysis about this method.
%%%%%%%%%%%%%%%%%

\subsection{S3VM-\emph{p}}\label{sec:graph}

The second baseline approach is motivated by confidence estimation in graph-based methods, e.g.,
\cite{zhu2003}, where the confidence can be naturally regarded as a risk measurement of unlabeled
data.

Formally, to estimate the confidence of unlabeled data, let $\bF^l =[(\y_l+1)/2,(1-\y_l)/2] \in \{0,1\}^{l \times 2}$ be the label matrix for labeled data where $\y_l = [y_1,\ldots,y_l]' \in \{\pm 1\}^{l \times 1} $ is the label vector. Let $\W = [w_{ij}] \in \mathcal{R}^{(l+u) \times (l+u)}$ be the weight matrix of training data and $\L$ is the
laplacian of $\W$, i.e., $\L= \bD-\W$ where $\bD = diag(d_{i})$ is a diagonal matrix with entries $d_i=\sum_{j}w_{ij}$. Then, the predictions of unlabeled data can be obtained by \cite{zhu2003}
\begin{equation}\label{eq:lap}
\bF^u = \L_{u,u}^{-1}\W_{u,l}\bF^l,
\end{equation}
where $\L_{u,u}$ is the sub-matrix of $\L$ with respect to the block of unlabeled data, while $\W_{u,l}$ is the sub-matrix of $\W$ with respect to the block between labeled and unlabeled data. Then, assign each point $\x_i$ with the label $y_{LabPo}(\x_i)=\sgn(\bF^{u}_{i-l,1}-\bF^{u}_{i-l,2})$ and the confidence $h_i = |\bF^{u}_{i-l,1}-\bF^{u}_{i-l,2}|$. After confidence estimation, similar to S3VM-\emph{c}, we consider the risk of unlabeled data by two factors, i.e., \emph{bias} and \emph{confidence}. If S3VM obtains the same bias of label propagation and the confidence is high enough, we use the S3VM prediction, and otherwise we take SVM prediction.

Algorithm~\ref{alg:graph} gives the S3VM-\emph{p} method and Figure~\ref{fig:illustration}(e) illustrates the intuition of S3VM-\emph{p}. As can be seen, the correct predictions of S3VM on groups $\{2,3\}$ are inherited by S3VM-\emph{p}, while the wrong predictions of S3VM on groups $\{7,8,9,10\}$ are avoided.

%\begin{figure}[htbp]
%\begin{center}
%\begin{tabular}{c}
%\subfigure{\psfig{figure=S3VM-\emph{us}-p1.eps, width = 1.5in}} \;\;\;\;\;\;
%\subfigure{\psfig{figure=S3VM-\emph{us}-p2.eps, width = 1.5in}}
%\end{tabular}
%\end{center}
%\vspace{-.2in} \caption{Illustration of the S3VM-\emph{p} method.} \label{fig:s3vm-p}
%\end{figure}

\begin{algorithm}[t]\small
\caption{~~S3VM-\emph{p}} \textbf{Input}: $y_{SVM}$, $y_{S3VM}$, $\DD$, $\W$ and parameter $\eta$
\begin{algorithmic}[1]
\STATE Perform label propagation (e.g., \cite{zhu2003}) with $\W$, obtain the predicted label
$y_{lp}(\x_i)$ and confidence $h_i$ for each unlabeled instance $\x_i$, $i=l+1,\ldots,l+u$.
\STATE Update $\h$ according to
\[h_i = y_{S3VM}(\x_i)y_{lp}(\x_i)h_i, i=l+1,\ldots,l+u. \]
Let $c$ denote the number of nonnegative entries in $\h$.
\STATE Sort $\h$, pick up the top-$\min\{\eta u,c\}$ values and use the predictions of S3VM for the corresponding unlabeled instances, otherwise use the predictions of SVM.
\end{algorithmic}\label{alg:graph}
\end{algorithm}

\begin{figure*}[htbp]
\centering
\begin{minipage}[c]{1.6 in}
\centering
\includegraphics[width =1.6 in]{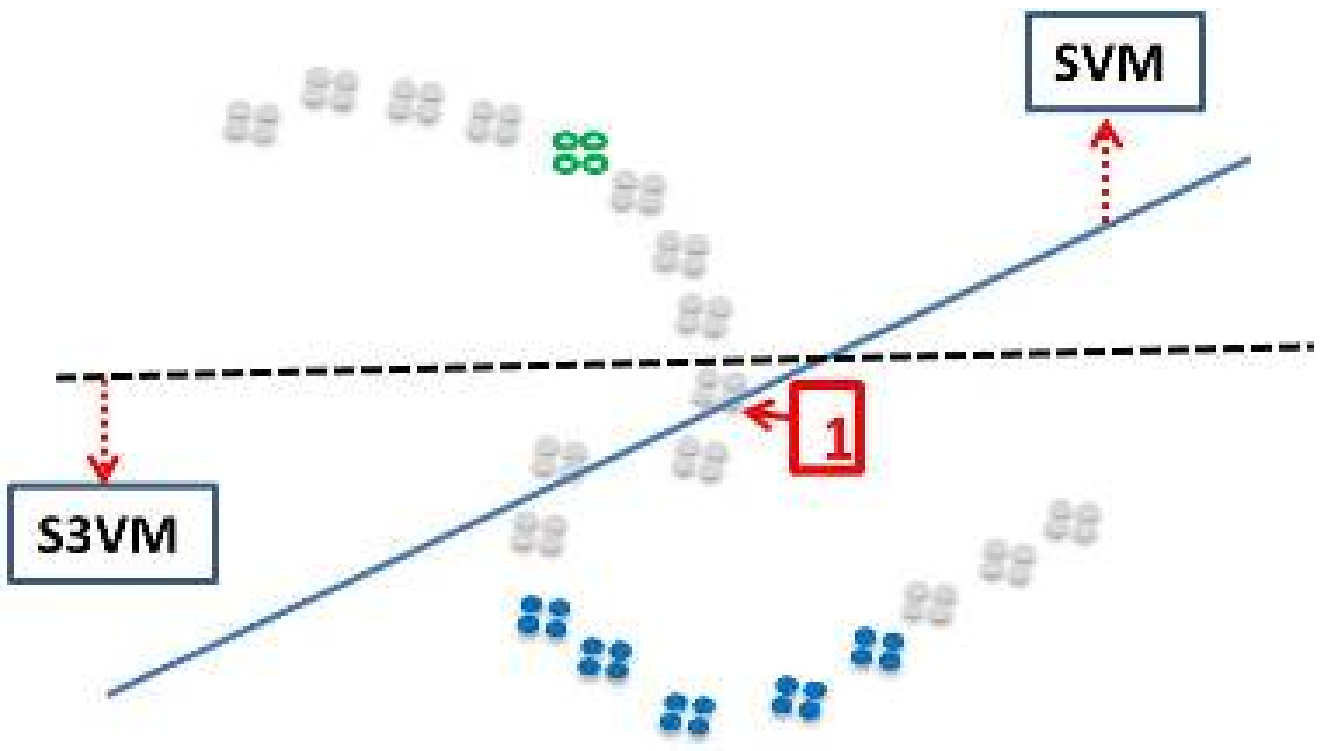}
\centering \mbox{\footnotesize (a)}
\end{minipage}
\centering
\begin{minipage}[c]{1.6 in}
\centering
\includegraphics[width =1.6 in]{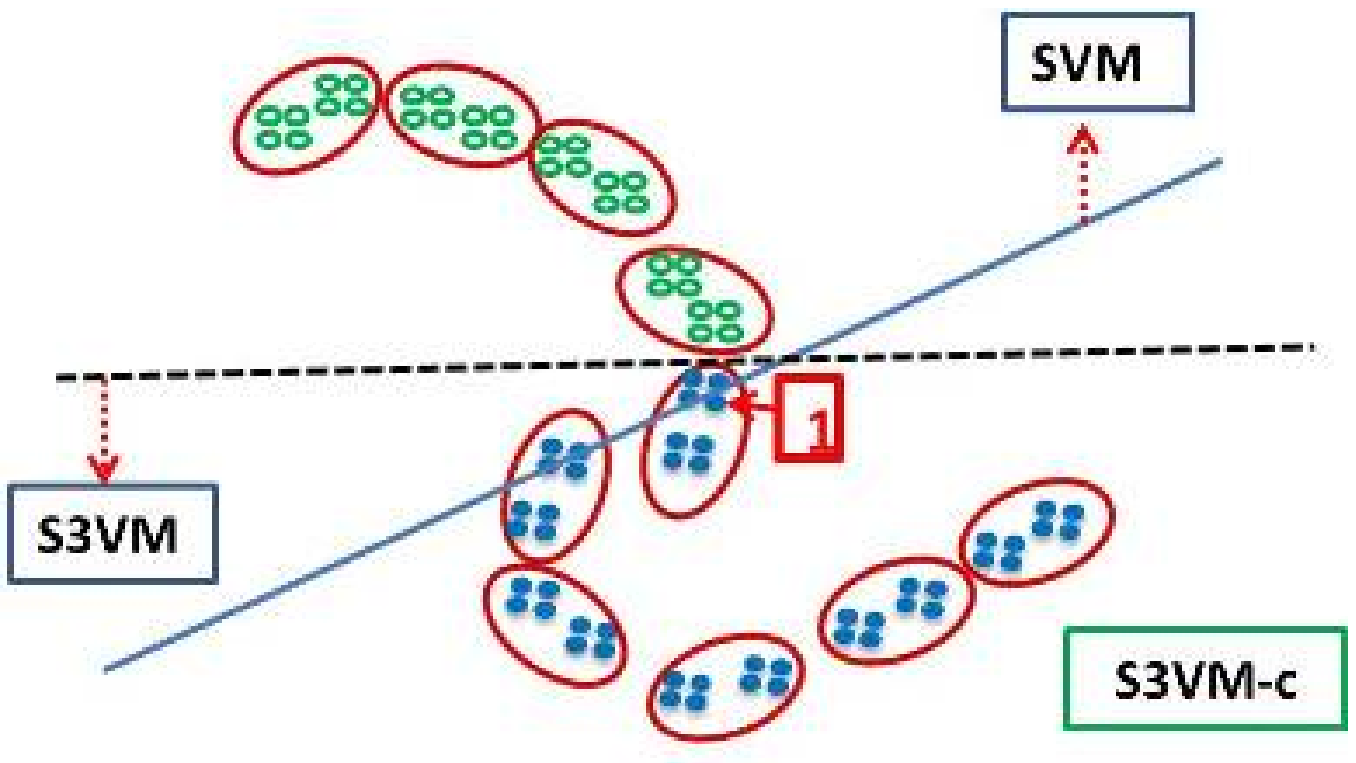}
\centering \mbox{\footnotesize (b)}
\end{minipage} \\
\begin{minipage}[c]{1.6 in}
\centering
\includegraphics[width =1.6 in]{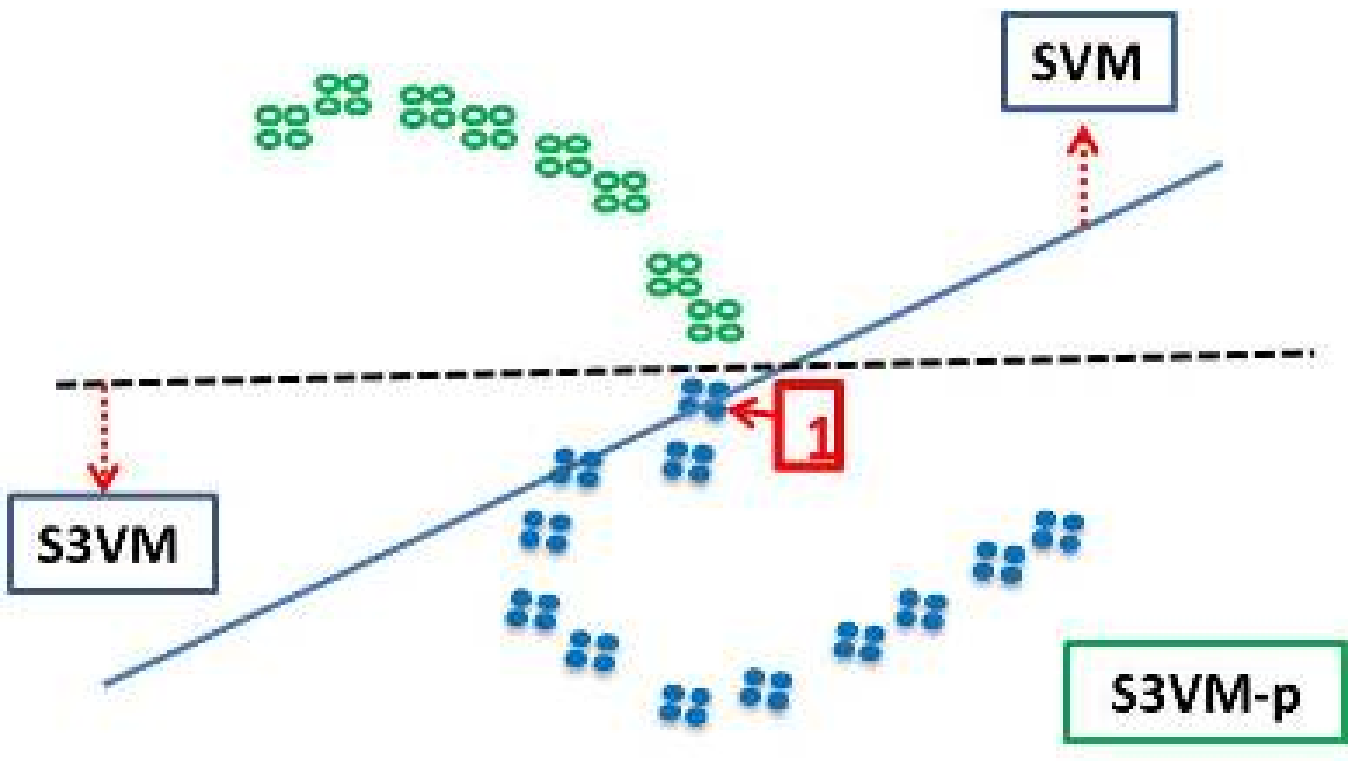}
\centering \mbox{\footnotesize (c)}
\end{minipage}
\begin{minipage}[c]{1.6 in}
\centering
\includegraphics[width =1.6 in]{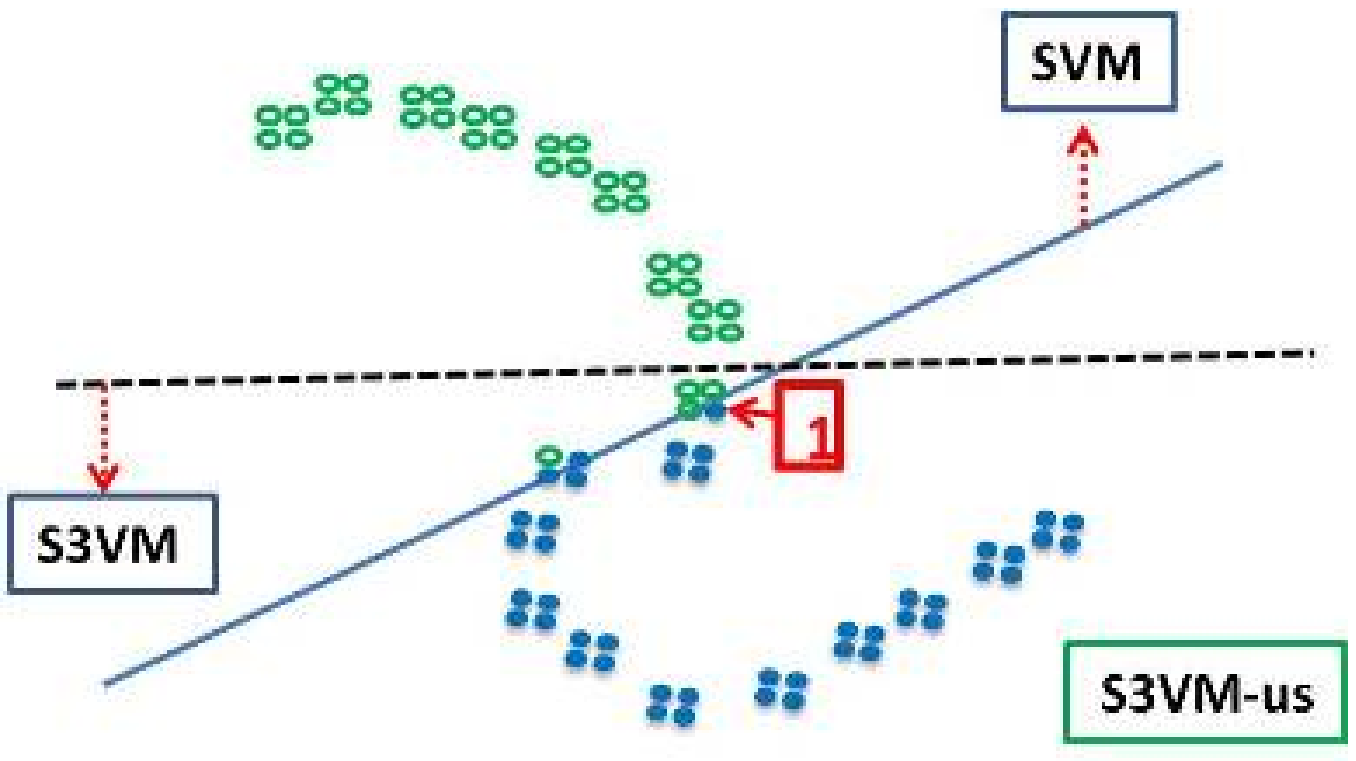}
\centering \mbox{\footnotesize (d)}
\end{minipage}
\caption{\small{Illustration with artificial two-moon data when S3VM degenerates performance. (a) Labeled data (empty and filled circles) and unlabeled data (gray points). The blocked number highlight a group of four unlabeled instances. Classification results of (b) S3VM-\emph{c}, where each circle presents a cluster; (c) S3VM-\emph{p}; (d) Our proposed
S3VM-\emph{us}.}}\label{fig:illustration-2}
\end{figure*}

\section{Our Proposed Method} \label{sec:hiecluster}

%%%%%%%%%%%%%%%%%
% What is the main motivation of using Hierarchical clustering? Alleviate the problem of graph construction
% and give some intuitive interpretations of hierarchical clustering method. How to use hierarchical clustering
% to design the concrete algorithm? Add some discussion about this method.
%%%%%%%%%%%%%%%%%%

\subsection{Deficiencies of S3VM-\emph{c} and S3VM-\emph{p}}

%By excluding some risky predictions of S3VM, both S3VM-\emph{c} and S3VM-\emph{p} methods can
%reduce the chances of degenerating performance and thus towards safe S3VMs. However, they both
%suffer from some difficulties.

S3VM-\emph{c} and S3VM-\emph{p} are capable of reducing the chances of performance degeneration by using unlabeled data, however, they both suffer from some deficiencies. For S3VM-\emph{c}, it works in a local manner and the relation between clusters are never considered, leading to the unexploitation of some helpful unlabeled instances, e.g., unlabeled instances in groups $\{2,3\}$ in Figure~\ref{fig:illustration-2}(d). For S3VM-\emph{p}, as stated in \cite{wang2008graph}, the confidence estimated by label propagation approach might be incorrect if the label initialization is highly imbalanced, leading to the unexploitation of some useful unlabeled instances, e.g., groups
$\{4,5\}$ in Figure~\ref{fig:illustration-2}(e).

Moreover, both S3VM-\emph{c} and S3VM-\emph{p} heavily rely on the predictions of S3VM, which might become a serious issue especially when S3VM obtains degenerated performance. Figures~\ref{fig:illustration-2}(b) and \ref{fig:illustration-2}(c) illustrate the behaviors of S3VM-\emph{c} and S3VM-\emph{p} when S3VM degenerates  performance. Both S3VM-\emph{c} and S3VM-\emph{p} erroneously inherit the wrong predictions of S3VM of group 1.

\subsection{S3VM-\emph{us}}

The deficiencies of S3VM-\emph{c} and S3VM-\emph{p} suggest to take into account of cluster relation and make the method insensitive to label initialization. This motivates us to use hierarchical clustering \cite{jain1988algorithms}, leading to our proposed method S3VM-\emph{us}.

Hierarchical clustering works in a greedy and iterative manner. It first initials each singe instance as a cluster and then at each step, it merges two clusters with the shortest distance among all pairs of clusters. In this step, the cluster relation is considered and moreover, since hierarchical clustering works in an unsupervised setting, it does not suffer from the label initialization problem.

Suppose $p_{i}$ and $n_{i}$ are the lengths of paths from the instance $\x_i$ to its nearest positive and negative labeled instances, respectively, in hierarchical clustering. We simply take the difference between $p_i$ and
$n_i$ as an estimation of the confidence on the unlabeled instance $\x_i$. Intuitively, the larger the difference between $p_i$ and $n_i$, the higher the confidence on labeling $\x_i$.

%To further reduce the chance of performance degeneration, alleviate the cases where SSL method achieves degenerated performance, SSL method is first compared with inductive method, if SSL method does not perform better than
%inductive method on the unlabeled instances, we simply output the inductive method instead.

Algorithm~\ref{alg:hie-cluster} gives the S3VM-\emph{us} method and Figures~\ref{fig:illustration}(f) and \ref{fig:illustration-2} illustrate the intuition of S3VM-\emph{us}. As can be seen, the wrong predictions of S3VM on groups $\{7,8,9,10\}$ are avoided by S3VM-\emph{us}, the correct predictions of S3VM on groups
$\{2,3,4,5\}$ are inherited, and S3VM-\emph{us} does not erroneously inherit the wrong predictions of S3VM
on group 1 in Figure 2.

%Thus, unlike S3VM-\emph{p}, S3VM-\emph{us} is more likely to correctly predict the points in groups 4-5 in
%Fig.\ref{fig:illustration}(f).
\vspace{-1mm}
\begin{algorithm}[t]\small
\caption{~~S3VM-\emph{us}} \textbf{Input}: $y_{SVM}$, $y_{S3VM}$, $\DD$ and parameter $\epsilon$
\begin{algorithmic}[1]
\STATE Let $\mathcal{S}$ be a set of the unlabeled data $\x$ such that $y_{SVM}(\x) \neq
y_{S3VM}(\x)$.
\STATE Perform hierarchical clustering, e.g., single linkage method
\cite{jain1988algorithms}.
\STATE For each unlabeled instance $\x_i \in \mathcal{S}$, calculate $p_i$ and $n_i$, that is, the length of the paths from $\x_i$ to its nearest positive and negative labeled instances, respectively. Denote $t_i = (n_i-p_i)$.
\STATE Let $\mathcal{B}$ be the set of unlabeled instances $\x_i$ in $\mathcal{S}$ satisfying $|t_i| \geq \epsilon|l+u|$.
\STATE If $\sum_{\x_i \in \mathcal{B}} y_{S3VM}(\x_i)t_i  \geq \sum_{\x_i \in \mathcal{B}} y_{SVM}(\x_i) t_i$, predict the unlabeled instances in $\mathcal{B}$ by S3VM and otherwise by SVM.
\STATE Predict the unlabeled data $\x \not\in \mathcal{B}$ by SVM.
\end{algorithmic}\label{alg:hie-cluster}
\end{algorithm}

\section{Experiments}
\label{sec:expt}

\subsection{Settings}

We evaluate S3VM-\emph{us} on a broad range of data sets including
the semi-supervised learning benchmark data sets in
\cite{chapelle2006ssl} and sixteen UCI data
sets\footnote{http://archive.ics.uci.edu/ml/}. The benchmark data
sets are {\small\sf g241c}, {\small\sf g241d}, {\small\sf Digit1},
{\small\sf USPS}, {\small\sf TEXT} and {\small\sf BCI}. For each
data, the
archive\footnote{{http://www.kyb.tuebingen.mpg.de/ssl-book/}}
provides two data sets with one using 10 labeled examples and the
other using 100 labeled examples. As for UCI data sets, we randomly
select 10 and 100 examples to be used as labeled examples,
respectively, and use the remaining data as unlabeled data. The
experiments are repeated for 30 times and the average accuracies and
standard deviations are recorded. It is worth noting that in
semi-supervised learning, labeled examples are often too few to
afford a valid cross validation, and therefore hold-out tests are
usually used for the evaluation.

%\begin{table}[ht]\small
%\caption{Data sets used in the experiments. }\smallskip\label{table:uci}
%\begin{center}
%\begin{tabular}{lccccc}
%\hline
%ID & Data &  \# Instances & \# Features \\
%1 & \textsf{BCI} & 400 & 117 \\
%2 & \textsf{g241c} & 1500 & 241 \\
%3 & \textsf{g241d} & 1500 & 241 \\
%4 & \textsf{Digit1} & 1500 & 241 \\
%5 & \textsf{USPS} & 1500 & 241 \\
%6 & \textsf{Text} & 1500 & 11960 \\
%7  & \textsf{house}   & 232 & 16 \\
%8  & \textsf{heart} & 270 & 9 \\
%9  & \textsf{heart-statlog} & 270 & 13\\
%10  & \textsf{ionosphere} & 351 & 33\\
%11  & \textsf{vehicle}  & 435 & 16 \\
%12  & \textsf{house-votes} & 435 & 16\\
%13  & \textsf{clean1} & 476 &  166\\
%14  & \textsf{wdbc}  & 569 & 14 \\
%15  & \textsf{isolet} & 600 & 51 \\
%16 & \textsf{breastw} & 683 & 9\\
%17 & \textsf{australian} & 690 & 15 \\
%18 & \textsf{diabetes} & 768 & 8\\
%19 & \textsf{german} & 1000 &  59 \\
%20 & \textsf{optdigits} & 1143 & 42 \\
%21 & \textsf{ethn} & 2630 & 30 \\
%22 & \textsf{sat} & 3041 & 36 \\
%\hline
%\end{tabular}
%\end{center}
%\end{table}

%%%%%%%%%%%%%%%%%%%%%%%%%%%%%%%
%% Short discription of the data sets and parameters setting.
%% Show the results and give some conclusive words
%%%%%%%%%%%%%%%%%%%%%%%%%%%%%%%

%Similar to Sec 4.1, SVM (using labeled data only) and TSVM are performed for comparison. As for ten
%labeled examples, parameter $C$ of SVM is set to 1. For fifty labeled examples and one hundred
%labeled examples, the parameter $C$ and the width of gaussian kernel are selected by cross
%validation. Other parameters are set as the same as in Sec 4.1.

In addition to S3VM-\emph{c} and S3VM-\emph{p}, we compare with inductive SVM and TSVM\footnote{{http://svmlight. joachims.org/}} \cite{Joachims1999}. Both linear and Gaussian kernels are used. For the benchmark data sets, we follow the setup in \cite{chapelle2006ssl}. Specifically, for the case of 10 labeled examples, the parameter $C$ for SVM is fixed to $m/\sum_{i=1}^{m}{\|\x_i\|^2}$ where $m=l+u$ is the size of data set and the Gaussian kernel width is set to $\delta$, i.e., the average distance between instances. For the case of $100$ labeled examples, $C$ is fixed to 100 and the Gaussian kernel width is selected from $\{0.25 \delta, 0.5 \delta, \delta, 2 \delta, 4 \delta\}$ by cross validation. On UCI data sets, the parameter $C$ is fixed to 1 and the Gaussian kernel width is set to $\delta$ for 10 labeled examples. For 100 label examples, the parameter $C$ is selected from $\{0.1,1,10,100\}$ and the Gaussian kernel width is selected from $\{0.25 \delta, 0.5 \delta, \delta, 2 \delta, 4 \delta\}$ by cross validation. For S3VM-\emph{c}, the cluster number $k$ is fixed to 50; for S3VM-\emph{p}, the weighted matrix is constructed via Gaussian distance and the parameter $\eta$ is fixed to 0.1; for S3VM-\emph{us}, the parameter $\epsilon$ is fixed to 0.1.

%%%%%%%%%%%%%%%%%%%%%%%%%%%%%%%%
\subsection{Results}\label{sec:benchmark}
%%%%%%%%%%%%%%%%%%%%%%%%%%%%%%%
%% Short discription of the UCI data sets and parameters setting.
%% Show the results and give some conclusive words

\begin{table*}[htbp]
\caption{\small{Accuracy (mean $ \pm$ std.) on 10 labeled examples. `SVM' denotes inductive SVM
which uses labeled data only. For the semi-supervised methods (TSVM, S3VM-\emph{c}, S3VM-\emph{p}
and S3VM-\emph{us}), if the performance is significantly better/worse than SVM, the corresponding
entries are bolded/underlined (paired $t$-tests at 95\% significance level). The win/tie/loss
counts with the fewest losses are bolded.}}\vspace{+2mm}
\smallskip \label{table:uci-result-1}
\begin{center}\scriptsize
\begin{tabular}{@{\!\!}ll@{\!\!}|c|cccc@{\!}}
\hline
 & Data & SVM & TSVM &  S3VM-\emph{c} & S3VM-\emph{p} & S3VM-\emph{us} \\
& & ( linear / gaussian ) & ( linear / gaussian ) & ( linear / gaussian ) & ( linear / gaussian ) & ( linear / gaussian ) \\
\hline
& \textsf{BCI} & 50.7$\pm$1.5 / 52.7$\pm$2.7 & \underline{49.3$\pm$2.8} / 51.4$\pm$2.7 & 50.2$\pm$2.0 / 52.2$\pm$2.6 & 50.6$\pm$1.6 / 52.6$\pm$2.7 & 50.9$\pm$1.6 / 52.6$\pm$2.7 \\
& \textsf{g241c} & 53.2$\pm$4.8 / 53.0$\pm$4.5 & \textbf{78.9$\pm$4.7} / \textbf{78.5$\pm$5.0} & 55.2$\pm$8.3 / 55.3$\pm$8.8 & \textbf{53.9$\pm$5.8} / \textbf{53.6$\pm$5.3} & 53.5$\pm$4.8 / 53.2$\pm$4.5 \\
& \textsf{g241d} & 54.4$\pm$5.4 / 54.5$\pm$5.2 & 53.6$\pm$7.8 / 53.2$\pm$6.5 & 53.8$\pm$5.4 / 53.6$\pm$5.0 & \underline{54.1$\pm$5.3} / \underline{54.0$\pm$5.2} & 54.4$\pm$5.3 / 54.4$\pm$5.2 \\
& \textsf{digit1} & 55.4$\pm$10.9 / 75.0$\pm$7.9 & \textbf{79.4$\pm$1.1} / \textbf{81.5$\pm$3.1} & 56.1$\pm$12.2 / \textbf{77.3$\pm$8.2} & 56.2$\pm$12.2 / 75.0$\pm$8.1 & \textbf{58.1$\pm$9.6} / 75.1$\pm$7.8 \\
& \textsf{USPS} & 80.0$\pm$0.1 / 80.7$\pm$1.8 & \underline{69.4$\pm$1.2} / \underline{73.0$\pm$2.6} & 80.0$\pm$0.1 / 80.4$\pm$2.5 & 80.0$\pm$0.1 / 80.5$\pm$2.1 & 80.0$\pm$0.1 / 80.7$\pm$1.8 \\
& \textsf{Text} & 54.7$\pm$6.3 / 54.6$\pm$6.3 & \textbf{71.4$\pm$11.7} / \textbf{71.2$\pm$11.4} & \textbf{56.8$\pm$8.8} / \textbf{56.5$\pm$8.7} & \textbf{55.3$\pm$6.6} / \textbf{55.2$\pm$6.8} & 58.0$\pm$9.0 / 57.8$\pm$8.9 \\
& \textsf{house} & 90.0$\pm$6.0 / 84.8$\pm$11.8 & \underline{84.6$\pm$8.0} / 84.7$\pm$6.9 & 89.8$\pm$6.2 / 84.8$\pm$11.9 & \underline{89.5$\pm$6.0} / \underline{84.5$\pm$11.8} & 90.1$\pm$6.1 / 85.4$\pm$11.4 \\
& \textsf{heart} & 58.8$\pm$10.5 / 63.9$\pm$11.6 & \textbf{72.4$\pm$12.6} / \textbf{72.6$\pm$10.4} & \textbf{59.0$\pm$10.8} / \textbf{64.4$\pm$11.6} & 58.6$\pm$10.6 / 63.8$\pm$11.7 & \textbf{61.9$\pm$9.7} / 65.1$\pm$11.0 \\
& \textsf{heart-statlog} & 74.6$\pm$4.8 / 69.9$\pm$10.1 & 74.9$\pm$6.6 / \textbf{73.9$\pm$5.9} & 74.5$\pm$5.2 / 70.1$\pm$10.2 & 74.5$\pm$4.9 / 70.0$\pm$10.2 & 74.2$\pm$5.4 / 71.7$\pm$6.9 \\
& \textsf{ionosphere} & 70.4$\pm$8.7 / 65.8$\pm$9.8 & 72.0$\pm$10.5 / \textbf{76.1$\pm$8.2} & \textbf{70.9$\pm$9.0} / \textbf{66.1$\pm$9.9} & 70.4$\pm$8.7 / 66.0$\pm$9.7 & 70.7$\pm$8.3 / 67.4$\pm$6.7 \\
& \textsf{vehicle} & 73.2$\pm$8.9 / 58.3$\pm$9.5 & 72.1$\pm$9.4 / \textbf{63.2$\pm$7.8} & 73.5$\pm$9.4 / 58.4$\pm$9.6 & \underline{72.6$\pm$9.1} / \underline{58.0$\pm$9.5} & \textbf{74.5$\pm$9.3} / \textbf{64.2$\pm$9.1} \\
& \textsf{house-votes} & 85.5$\pm$7.0 / 79.7$\pm$10.7 & 83.8$\pm$6.1 / \textbf{84.0$\pm$5.3} & 85.7$\pm$7.0 / 80.1$\pm$10.6 & 85.3$\pm$6.9 / 79.7$\pm$10.7 & 86.0$\pm$5.7 / \textbf{84.3$\pm$6.1} \\
& \textsf{wdbc} & 65.6$\pm$7.5 / 73.8$\pm$10.3 & \textbf{90.0$\pm$6.1} / \textbf{88.9$\pm$3.7} & 65.7$\pm$7.8 / \textbf{74.9$\pm$10.9} & \textbf{66.1$\pm$8.0} / 73.9$\pm$10.5 & \textbf{65.8$\pm$7.5} / \textbf{73.9$\pm$10.3} \\
& \textsf{clean1} & 58.2$\pm$4.2 / 53.5$\pm$6.2 & 57.0$\pm$5.1 / 53.3$\pm$4.8 & 57.8$\pm$4.4 / 53.3$\pm$6.2 & \textbf{58.5$\pm$4.2} / 53.3$\pm$6.3 & 58.2$\pm$4.2 / \textbf{55.0$\pm$8.1} \\
& \textsf{isolet} & 93.8$\pm$4.3 / 82.0$\pm$15.7 & \underline{84.2$\pm$10.9} / \textbf{86.7$\pm$9.5} & \textbf{94.5$\pm$5.1} / \textbf{83.2$\pm$16.0} & \underline{93.0$\pm$4.7} / \underline{81.7$\pm$15.7} & 93.7$\pm$4.3 / \textbf{84.1$\pm$12.6} \\
& \textsf{breastw} & 93.9$\pm$4.8 / 92.3$\pm$10.1 & \underline{89.2$\pm$8.6} / 88.9$\pm$8.8 & 94.2$\pm$4.9 / 92.4$\pm$10.0 & 93.9$\pm$4.9 / 92.2$\pm$10.0 & 93.6$\pm$5.4 / 92.4$\pm$9.9 \\
& \textsf{australian} & 70.4$\pm$9.2 / 60.3$\pm$8.4 & 69.6$\pm$11.9 / \textbf{68.6$\pm$11.4} & 70.1$\pm$9.8 / 60.4$\pm$8.3 & 70.5$\pm$9.4 / 60.5$\pm$8.8 & 70.3$\pm$9.2 / 60.8$\pm$7.9 \\
& \textsf{diabetes} & 63.3$\pm$6.9 / 66.3$\pm$3.5 & 63.4$\pm$7.6 / 65.8$\pm$4.6 & 63.2$\pm$6.8 / \underline{65.9$\pm$3.0} & 63.4$\pm$6.6 / 66.2$\pm$3.4 & 63.3$\pm$6.9 / 66.3$\pm$3.5 \\
& \textsf{german} & 65.2$\pm$4.9 / 65.1$\pm$12.0 & 63.7$\pm$5.6 / 63.5$\pm$5.1 & 65.6$\pm$4.7 / 65.1$\pm$11.8 & \textbf{65.6$\pm$4.8} / 65.1$\pm$11.9 & 65.2$\pm$5.0 / 65.3$\pm$11.6 \\
& \textsf{optdigits} & 96.1$\pm$3.2 / 92.8$\pm$9.6 & \underline{89.8$\pm$9.2} / 91.4$\pm$7.6 & \textbf{96.6$\pm$3.1} / \textbf{93.6$\pm$9.9} & \underline{95.6$\pm$3.0} / \underline{92.4$\pm$9.8} & 96.9$\pm$2.5 / 94.9$\pm$5.8 \\
& \textsf{ethn} & 56.5$\pm$8.8 / 58.5$\pm$10.2 & \textbf{64.2$\pm$13.5} / \textbf{68.1$\pm$14.5} & 56.5$\pm$8.6 / \textbf{59.4$\pm$11.6} & 56.8$\pm$9.1 / 58.6$\pm$10.7 & \textbf{59.8$\pm$10.7} / \textbf{61.8$\pm$11.3} \\
& \textsf{sat} & 95.8$\pm$4.1 / 87.5$\pm$10.9 & \underline{85.5$\pm$11.4} / 86.5$\pm$10.8 & \textbf{96.3$\pm$4.1} / 87.7$\pm$11.2 & \underline{94.8$\pm$4.2} / \underline{86.9$\pm$10.8} & 96.4$\pm$3.9 / \textbf{90.7$\pm$8.1} \\
\hline
&{Aver. Acc.}  &70.9 / 69.3&73.5 / {73.8}&71.2 / 69.8&70.9 / 69.3&{71.6} / 70.8 \\
\hline
&\multicolumn{2}{@{\;}c@{\;}|}{SVM vs. Semi-Supervised: W/T/L} & 18/18/8 & 14/29/1 & 7/25/12 & \textbf{12/32/0}\\
\hline  \noalign{\smallskip} \hline
\end{tabular}
\end{center}
\end{table*}

\begin{table*}[htbp]
\caption{\small{Accuracy (mean $ \pm$ std.) on 100 labeled examples. `SVM' denotes inductive SVM
which uses labeled data only. For the semi-supervised methods (TSVM, S3VM-\emph{c}, S3VM-\emph{p}
and S3VM-\emph{us}), if the performance is significantly better/worse than SVM, the corresponding
entries are bolded/underlined (paired $t$-tests at 95\% significance level). The win/tie/loss
counts with the fewest losses are bolded.}}\vspace{+2mm}
\smallskip \label{table:uci-result-2}
\begin{center}\scriptsize
\begin{tabular}{@{\!\!}ll@{\!\!}|c|cccc@{\!}}
\hline
& Data & SVM & TSVM &  S3VM-\emph{c} & S3VM-\emph{p} & S3VM-\emph{us} \\
& & ( linear / gaussian ) & ( linear / gaussian ) & ( linear / gaussian ) & ( linear / gaussian ) & ( linear / gaussian ) \\
\hline
& \textsf{BCI} & 61.1$\pm$2.6 / 65.9$\pm$3.1 & \underline{56.4$\pm$2.8} / 65.6$\pm$2.5 & \underline{58.3$\pm$2.6} / 65.6$\pm$3.0 & \underline{60.3$\pm$2.5} / 65.8$\pm$3.0 & 61.0$\pm$2.7 / 65.8$\pm$3.1 \\
& \textsf{g241c} & 76.3$\pm$2.0 / 76.6$\pm$2.1 & \textbf{81.7$\pm$1.6} / \textbf{82.1$\pm$1.2} & \textbf{79.3$\pm$1.7} / \textbf{79.6$\pm$1.8} & \textbf{77.2$\pm$2.1} / \textbf{77.1$\pm$2.0} & 76.3$\pm$2.0 / 76.6$\pm$2.1 \\
& \textsf{g241d} & 74.2$\pm$1.9 / 75.4$\pm$1.8 & 76.1$\pm$8.5 / 77.9$\pm$7.4 & \textbf{77.4$\pm$3.5} / \textbf{78.5$\pm$3.3} & \textbf{74.8$\pm$2.3} / 75.7$\pm$2.2 & 74.2$\pm$1.9 / 75.4$\pm$1.8 \\
& \textsf{digit1} & 50.3$\pm$1.2 / 94.0$\pm$1.4 & \textbf{81.9$\pm$3.0} / 94.0$\pm$2.0 & 50.3$\pm$1.2 / \textbf{95.0$\pm$1.5} & 50.3$\pm$1.2 / 94.1$\pm$1.4 & \textbf{67.9$\pm$1.3} / 94.1$\pm$1.4 \\
& \textsf{USPS} & 80.0$\pm$0.2 / 91.7$\pm$1.1 & 78.8$\pm$2.0 / 90.9$\pm$1.4 & 80.0$\pm$0.2 / \textbf{92.5$\pm$1.0} & 80.0$\pm$0.2 / \underline{91.6$\pm$1.2} & 80.1$\pm$0.4 / 91.8$\pm$1.1 \\
& \textsf{Text} & 73.8$\pm$3.3 / 73.7$\pm$3.6 & \textbf{77.7$\pm$1.6} / \textbf{77.7$\pm$1.7} & \textbf{75.3$\pm$3.4} / \textbf{75.2$\pm$3.6} & \textbf{73.9$\pm$3.4} / \textbf{73.8$\pm$3.7} & 74.1$\pm$3.1 / \textbf{74.2$\pm$3.3} \\
& \textsf{house} & 95.7$\pm$2.0 / 95.6$\pm$1.6 & \underline{94.4$\pm$2.5} / 94.8$\pm$2.6 & 95.5$\pm$1.8 / 95.4$\pm$1.8 & 95.6$\pm$2.0 / 95.5$\pm$1.7 & 95.6$\pm$2.0 / 95.6$\pm$1.6 \\
& \textsf{heart} & 81.5$\pm$2.5 / 80.1$\pm$2.4 & \underline{80.7$\pm$3.1} / 79.5$\pm$2.9 & 81.1$\pm$3.0 / \underline{79.8$\pm$2.5} & 81.5$\pm$2.5 / 80.2$\pm$2.5 & 81.5$\pm$2.6 / 80.1$\pm$2.4 \\
& \textsf{heart-statlog} & 81.5$\pm$2.4 / 81.4$\pm$2.7 & 81.6$\pm$2.7 / \underline{79.0$\pm$4.5} & 81.2$\pm$2.2 / \underline{80.7$\pm$3.0} & 81.5$\pm$2.4 / \underline{81.2$\pm$2.7} & 81.5$\pm$2.4 / 81.3$\pm$2.7 \\
& \textsf{ionosphere} & 87.1$\pm$1.5 / 93.2$\pm$1.6 & \underline{85.6$\pm$2.1} / \underline{92.1$\pm$2.3} & \textbf{88.7$\pm$1.3} / 93.4$\pm$1.5 & 87.1$\pm$1.5 / 93.2$\pm$1.6 & 87.1$\pm$1.5 / 93.2$\pm$1.6 \\
& \textsf{vehicle} & 92.9$\pm$1.7 / 95.4$\pm$1.4 & \underline{91.6$\pm$2.5} / 95.4$\pm$2.3 & \textbf{93.3$\pm$1.6} / \textbf{95.9$\pm$1.3} & \underline{92.8$\pm$1.7} / 95.2$\pm$1.5 & \textbf{93.0$\pm$1.7} / \textbf{95.5$\pm$1.4} \\
& \textsf{house-votes} & 92.3$\pm$1.3 / 92.8$\pm$1.2 & 92.0$\pm$1.8 / 93.0$\pm$1.4 & 92.6$\pm$1.2 / 92.9$\pm$1.2 & 92.3$\pm$1.3 / 92.8$\pm$1.2 & 92.3$\pm$1.3 / 92.8$\pm$1.2 \\
& \textsf{clean1} & 73.0$\pm$2.7 / 80.6$\pm$3.0 & 73.2$\pm$3.1 / \underline{79.1$\pm$3.4} & \textbf{73.7$\pm$2.9} / \underline{79.9$\pm$2.9} & \textbf{73.2$\pm$2.6} / \underline{80.4$\pm$3.2} & 73.1$\pm$2.7 / 80.7$\pm$3.0 \\
& \textsf{wdbc} & 95.6$\pm$0.8 / 94.7$\pm$0.9 & \underline{94.3$\pm$2.3} / 94.1$\pm$2.4 & \textbf{95.8$\pm$0.7} / 94.9$\pm$0.9 & 95.6$\pm$0.8 / 94.7$\pm$0.9 & \textbf{95.6$\pm$0.8} / \textbf{94.8$\pm$0.9} \\
& \textsf{isolet} & 99.2$\pm$0.4 / 99.0$\pm$0.6 & \underline{95.9$\pm$3.1} / 98.2$\pm$2.3 & 99.2$\pm$0.4 / \textbf{99.2$\pm$0.5} & \underline{99.0$\pm$0.4} / 98.9$\pm$0.6 & 99.2$\pm$0.4 / \textbf{99.1$\pm$0.5} \\
& \textsf{breastw} & 96.4$\pm$0.4 / 96.7$\pm$0.4 & 96.9$\pm$1.9 / \textbf{97.1$\pm$0.5} & \textbf{96.6$\pm$0.4} / \textbf{96.9$\pm$0.4} & 96.3$\pm$0.4 / 96.7$\pm$0.4 & 96.4$\pm$0.4 / 96.7$\pm$0.4 \\
& \textsf{australian} & 83.8$\pm$1.6 / 84.9$\pm$1.7 & \underline{82.5$\pm$2.6} / 84.6$\pm$2.7 & 83.8$\pm$1.7 / 85.0$\pm$1.6 & 83.9$\pm$1.7 / 85.0$\pm$1.8 & 83.8$\pm$1.7 / 85.0$\pm$1.7 \\
& \textsf{diabetes} & 75.2$\pm$1.7 / 74.7$\pm$1.9 & \underline{72.3$\pm$2.3} / \underline{71.8$\pm$1.8} & \underline{74.9$\pm$1.7} / \underline{74.2$\pm$2.2} & 75.3$\pm$1.6 / 74.7$\pm$1.9 & 75.2$\pm$1.8 / 74.7$\pm$1.9 \\
& \textsf{german} & 67.1$\pm$2.4 / 72.0$\pm$1.5 & \underline{66.1$\pm$2.1} / \underline{65.9$\pm$3.4} & 67.1$\pm$2.2 / \underline{71.6$\pm$1.5} & \textbf{67.6$\pm$2.3} / 72.1$\pm$1.4 & 67.1$\pm$2.4 / 72.1$\pm$1.5 \\
& \textsf{optdigits} & 99.4$\pm$0.3 / 99.4$\pm$0.3 & \underline{95.9$\pm$3.7} / \underline{97.4$\pm$3.1} & 99.5$\pm$0.4 / \textbf{99.5$\pm$0.3} & \underline{99.2$\pm$0.4} / \underline{99.2$\pm$0.4} & 99.5$\pm$0.3 / 99.4$\pm$0.3 \\
& \textsf{ethn} & 91.6$\pm$1.6 / 93.4$\pm$1.2 & \textbf{92.6$\pm$2.3} / 93.4$\pm$3.0 & \textbf{93.9$\pm$1.6} / \textbf{95.0$\pm$1.2} & \textbf{91.9$\pm$1.5} / \underline{93.3$\pm$1.2} & \textbf{91.7$\pm$1.5} / 93.4$\pm$1.2 \\
& \textsf{sat} & 99.7$\pm$0.2 / 99.7$\pm$0.1 & \underline{96.4$\pm$2.8} / \underline{97.6$\pm$2.7} & \textbf{99.7$\pm$0.2} / \textbf{99.8$\pm$0.1} & \underline{99.5$\pm$0.3} / \underline{99.5$\pm$0.3} & 99.7$\pm$0.2 / 99.7$\pm$0.1 \\
\hline
&{Aver. Acc.}  &83.0 / {86.8}&83.9 / 86.4&{83.5 / 87.3}&83.1 / 86.8&83.9 / {86.9}\\
\hline
&\multicolumn{2}{@{\;}c@{\;}|}{SVM vs. Semi-Supervised: W/T/L} & 7/18/19 & 21/16/7 & 8/25/11 & \textbf{{8/36/0}}\\
\hline \hline
\end{tabular}
\end{center}
\end{table*}

The results are shown in Tables~\ref{table:uci-result-1} and~\ref{table:uci-result-2}. As can be
seen, the performance of S3VM-\emph{us} is competitive with TSVM. In terms of average accuracy,
TSVM performs slightly better (worse) than S3VM-\emph{us} on the case of 10 (100) labeled examples.
In terms of pairwise comparison, S3VM-\emph{us} performs better than TSVM on 13/12 and 14/16 cases
with linear/Gaussian kernel for 10 and 100 labeled examples, respectively. Note that in a number of
cases, TSVM has large performance improvement against inductive SVM, while the improvement of
S3VM-\emph{us} is smaller. This is not a surprise since S3VM-\emph{us} tries to improve performance
with the caution of avoiding performance degeneration.

Though TSVM has large improvement in a number of cases, it also has large performance degeneration
in cases. Indeed, as can be seen from Tables~\ref{table:uci-result-1} and~\ref{table:uci-result-2},
TSVM is significantly inferior to inductive SVM on 8/44, 19/44 cases for 10 and 100 labeled
examples, respectively. Both S3VM-\emph{c} and S3VM-\emph{p} are capable to reduce the times of
significant performance degeneration, while S3VM-\emph{us} does not significantly degenerate
performance in the experiments.

%These experimental results indicate although S3VM-\emph{us} is directly performed on the prediction
%of TSVM, it effectively selects useful unlabeled data to improve the performance while what is more
%important, it successfully excludes high risky unlabeled data such that the performance would never
%been degenerated.

\subsection{Parameter Influence}

\begin{figure}[t]
\centering

\includegraphics[width =3.5 in]{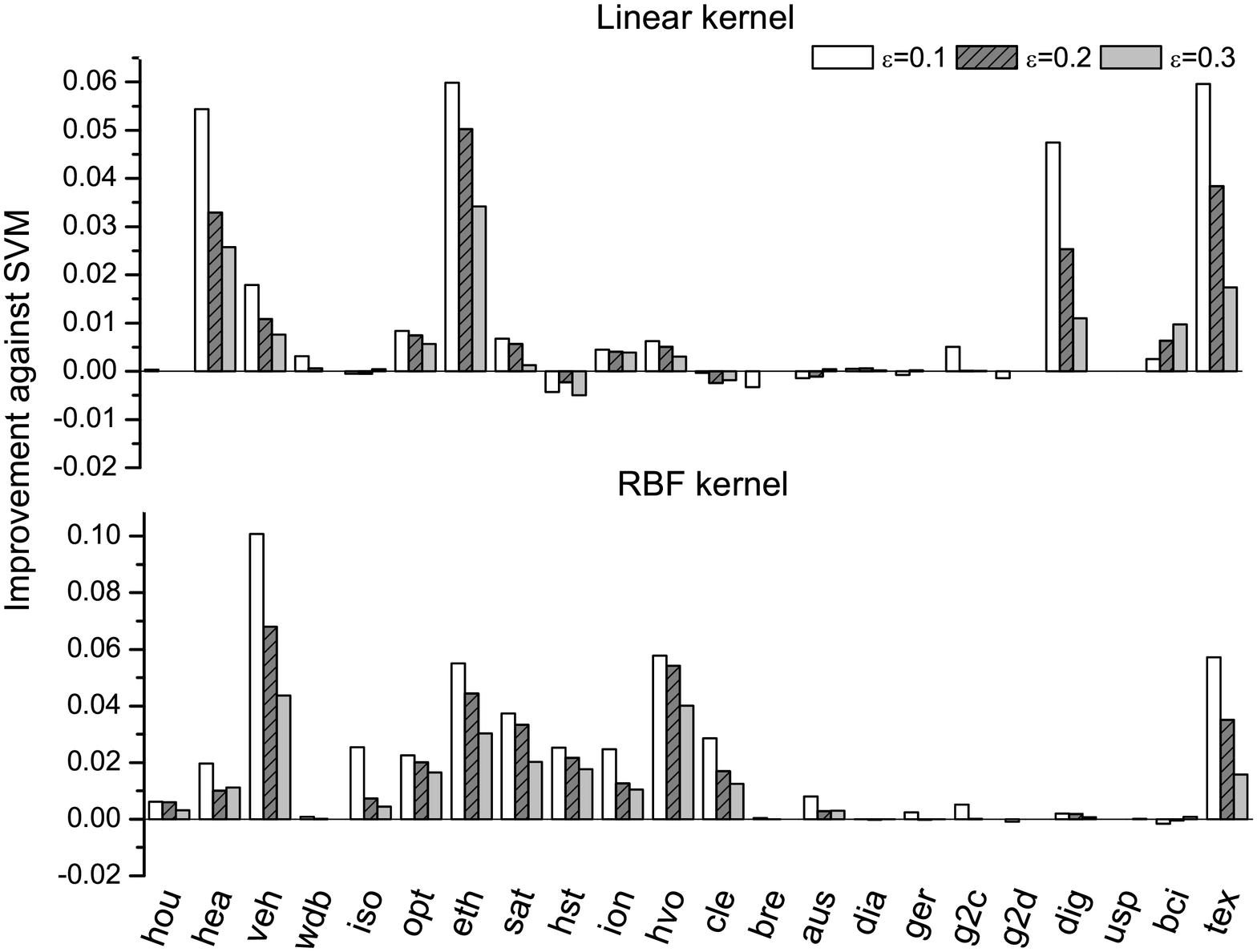}
\centering
\caption{Influence of the parameter $\epsilon$ on the
improvement of S3VM-\emph{us} against inductive
SVM.}\label{fig:S3VM-epsilon}
\end{figure}

S3VM-\emph{us} has a parameter $\epsilon$. To study the influence of $\epsilon$, we run experiments
by setting $\epsilon$ to different values (0.1, 0.2 and 0.3) with 10 labeled examples. The results
are plotted in Figure~\ref{fig:S3VM-epsilon}. It can be seen that the setting of $\epsilon$ has
influence on the improvement of S3VM-\emph{us} against inductive SVM. Whatever linear kernel or
gaussian kernel is used, the larger the value of $\epsilon$, the closer the performance of
S3VM-\emph{us} to SVM. It may be possible to increase the performance improvement by setting a
smaller $\epsilon$, however, this may increase the risk of performance degeneration.

%%%%%%%%%%%%%%%%%%%%%%%%%%%%%%%%%%%%%%%%%%%%%%%%%%%%%%%%%%%%%%%%%%%%%%%%
\vspace{+2mm}

\section{Conclusion}

In this paper we propose the S3VM-\emph{us} method. Rather than
simply predicting all unlabeled instances by semi-supervised
learner, S3VM-\emph{us} uses hierarchical clustering to help select
unlabeled instances to be predicted by semi-supervised learner and
predict the remaining unlabeled instances by inductive learner. In
this way, the risk of performance degeneration by using unlabeled
data is reduced. The effectiveness of S3VM-\emph{us} is validated by
empirical study.

The proposal in this paper is based on heuristics and theoretical
analysis is future work. It is worth noting that, along with
reducing the chance of performance degeneration, S3VM-\emph{us} also
reduces the possible performance gains from unlabeled data. In the
future it is desirable to develop really \textit{safe}
semi-supervised learning approaches which are able to improve
performance significantly but never degenerate performance by using
unlabeled data.

%\section*{Acknowledgments}
%
%We want to thank anonymous reviewers for helpful comments. This research was supported by the NSFC
%(61073097), JiangsuSF (BK2008018) and 973 Program (2010CB327903).

\small
\bibliographystyle{plain}
\bibliography{arxiv-s3vmus}

\begin{thebibliography}{10}

\bibitem{balcan2005person}
M.~F. Balcan, A.~Blum, P.~P. Choi, J.~Lafferty, B.~Pantano, M.~R. Rwebangira,
  and X.~Zhu.
\newblock {Person identification in webcam images: An application of
  semi-supervised learning}.
\newblock In {\em ICML Workshop on Learning with Partially Classified Training
  Data}, 2005.

\bibitem{ben2008does}
S.~Ben-David, T.~Lu, and D.~P{\'a}l.
\newblock {Does unlabeled data provably help? Worst-case analysis of the sample
  complexity of semi-supervised learning}.
\newblock In {\em COLT}, pages 33--44, 2008.

\bibitem{bennett1999sss}
K.~Bennett and A.~Demiriz.
\newblock Semi-supervised support vector machines.
\newblock In {\em {NIPS} 11}, pages 368--374. 1999.

\bibitem{blum2001}
A.~Blum and S.~Chawla.
\newblock {Learning from labeled and unlabeled data using graph mincuts}.
\newblock In {\em ICML}, pages 19--26, 2001.

\bibitem{blum1998}
A.~Blum and T.~Mitchell.
\newblock {Combining labeled and unlabeled data with co-training}.
\newblock In {\em COLT}, pages 92--100, 1998.

\bibitem{chapelle2006ssl}
O.~Chapelle, B.~Sch{\"o}lkopf, and A.~Zien, editors.
\newblock {\em Semi-Supervised Learning}.
\newblock MIT Press, Cambridge, MA, 2006.

\bibitem{Chapelle2008}
O.~Chapelle, V.~Sindhwani, and S.~S. Keerthi.
\newblock Optimization techniques for semi-supervised support vector machines.
\newblock {\em J. Mach. Learn. Res.}, 9:203--233, 2008.

\bibitem{Chapelle2005}
O.~Chapelle and A.~Zien.
\newblock Semi-supervised learning by low density separation.
\newblock In {\em AISTATS}, pages 57--64, 2005.

\bibitem{chawla2005learning}
N.~V. Chawla and G.~Karakoulas.
\newblock Learning from labeled and unlabeled data: An empirical study across
  techniques and domains.
\newblock {\em J. Artif. Intell. Res.}, 23:331--366, 2005.

\bibitem{collobert2006lst}
R.~Collobert, F.~Sinz, J.~Weston, and L.~Bottou.
\newblock Large scale transductive {SVMs}.
\newblock {\em J. Mach. Learn. Res.}, 7:1687--1712, 2006.

\bibitem{Cozman2003}
F.~G. Cozman, I.~Cohen, and M.~C. Cirelo.
\newblock Semi-supervised learning of mixture models.
\newblock In {\em ICML}, pages 99--106, 2003.

\bibitem{dasgupta2002pac}
S.~Dasgupta, M.~L. Littman, and D.~McAllester.
\newblock {PAC} generalization bounds for co-training.
\newblock In {\em {NIPS} 14}, pages 375--382. 2002.

\bibitem{jain1988algorithms}
A.K. Jain and R.C. Dubes.
\newblock {\em {Algorithms for Clustering Data}}.
\newblock Prentice Hall, Englewood Cliffs, NJ., 1988.

\bibitem{Jebara2009}
T.~Jebara, J.~Wang, and S.~F. Chang.
\newblock Graph construction and b-matching for semi-supervised learning.
\newblock In {\em ICML}, pages 441--448, 2009.

\bibitem{Joachims1999}
T.~Joachims.
\newblock Transductive inference for text classification using support vector
  machines.
\newblock In {\em ICML}, pages 200--209, 1999.

\bibitem{LaffertyNIPS2007}
J.~Lafferty and L.~Wasserman.
\newblock Statistical analysis of semi-supervised regression.
\newblock In {\em NIPS 20}, pages 801--808. 2008.

\bibitem{li2005setred}
M.~Li and Z.~H. Zhou.
\newblock {SETRED}: Self-training with editing.
\newblock In {\em PAKDD}, pages 611--621, 2005.

\bibitem{li2009means3vm}
Y.-F. Li, J.~T. Kwok, and Z.-H. Zhou.
\newblock Semi-supervised learning using label mean.
\newblock In {\em ICML}, pages 633--640, 2009.

\bibitem{Miller:Uyar1997}
D.~J. Miller and H.~S. Uyar.
\newblock A mixture of experts classifier with learning based on both labelled
  and unlabelled data.
\newblock In {\em NIPS 9}, pages 571--577. 1997.

\bibitem{nigam2000text}
K.~Nigam, A.~K. McCallum, S.~Thrun, and T.~Mitchell.
\newblock {Text classification from labeled and unlabeled documents using EM}.
\newblock {\em Mach. Learn.}, 39(2):103--134, 2000.

\bibitem{SinghNIPS2008}
A.~Singh, R.~Nowak, and X.~Zhu.
\newblock Unlabeled data: Now it helps, now it doesn't.
\newblock In {\em NIPS 21}, pages 1513--1520. 2009.

\bibitem{wang2008graph}
J.~Wang, T.~Jebara, and S.~F. Chang.
\newblock {Graph transduction via alternating minimization}.
\newblock In {\em ICML}, pages 1144--1151, 2008.

\bibitem{wang2007}
W.~Wang and Z.-H. Zhou.
\newblock Analyzing co-training style algorithms.
\newblock In {\em ECML}, pages 454--465, 2007.

\bibitem{Wang:Zhou2010}
W.~Wang and Z.-H. Zhou.
\newblock A new analysis of co-training.
\newblock In {\em ICML}, pages 1135--1142, 2010.

\bibitem{zhang2000}
T.~Zhang and F.~Oles.
\newblock {The value of unlabeled data for classification problems}.
\newblock In {\em ICML}, pages 1191--1198, 2000.

\bibitem{zhou2004}
D.~Zhou, O.~Bousquet, T.~N. Lal, J.~Weston, and B.~Scholkopf.
\newblock Learning with local and global consistency.
\newblock In {\em {NIPS} 16}, pages 595--602. 2004.

\bibitem{zhou2005tri}
Z.-H. Zhou and M.~Li.
\newblock Tri-training: Exploiting unlabeled data using three classifiers.
\newblock {\em IEEE Trans. Knowl. Data Eng.}, 17(11):1529--1541, 2005.

\bibitem{Zhou:Li2010}
Z.-H. Zhou and M.~Li.
\newblock Semi-supervised learning by disagreement.
\newblock {\em Knowl. Inf. Syst.}, 24(3):415--439, 2010.

\bibitem{zhu2007semi}
X.~Zhu.
\newblock Semi-supervised learning literature survey.
\newblock Technical Report 1530, Dept. Comp. Sci., Univ. Wisconsin-Madison,
  2006.

\bibitem{zhu2003}
X.~Zhu, Z.~Ghahramani, and J.~D. Lafferty.
\newblock Semi-supervised learning using gaussian fields and harmonic
  functions.
\newblock In {\em ICML}, pages 912--919, 2003.

\end{thebibliography}

\end{document}